\documentclass[10pt,twocolumn,letterpaper]{article}

\usepackage[pagenumbers]{cvpr} 

\usepackage{graphicx}
\usepackage{amsmath}
\usepackage{amssymb}
\usepackage{booktabs}
\usepackage{multirow}
\usepackage{pifont}
\usepackage[dvipsnames]{xcolor} 
\usepackage{threeparttable}
\usepackage{wrapfig}

\usepackage[pagebackref,breaklinks,colorlinks]{hyperref}

\usepackage[capitalize]{cleveref}
\crefname{section}{Sec.}{Secs.}
\Crefname{section}{Section}{Sections}
\Crefname{table}{Table}{Tables}
\crefname{table}{Tab.}{Tabs.}

\usepackage{amsmath,amsfonts,bm}

\def\eqref#1{equation~\ref{#1}}

\def\1{\bm{1}}

\DeclareMathAlphabet{\mathsfit}{\encodingdefault}{\sfdefault}{m}{sl}
\SetMathAlphabet{\mathsfit}{bold}{\encodingdefault}{\sfdefault}{bx}{n}

\newcommand{\R}{\mathbb{R}}

\newcommand\name{MSG-Transformer}
\newcommand\msg{\texttt{MSG}}
\newcommand{\spaceline}{\specialrule{0em}{1pt}{1pt}}
\newcommand{\xmark}{\ding{55}}

\def\cvprPaperID{2434} 
\def\confName{CVPR}
\def\confYear{2022}

\begin{document}

\title{\name{}: Exchanging Local Spatial Information by \\Manipulating Messenger Tokens}

\author{
Jiemin Fang$^{1,2}$ ,  Lingxi Xie$^{3}$ ,  Xinggang Wang$^2$\footnotemark[2] ,  Xiaopeng Zhang$^{3}$ ,  Wenyu Liu$^2$ ,  Qi Tian$^{3}$\\
$^1$Institute of Artificial Intelligence, Huazhong University of Science \& Technology\\
$^2$School of EIC, Huazhong University of Science \& Technology \; $^3$Huawei Inc.\\
\texttt{\small\{jaminfong, xgwang, liuwy\}@hust.edu.cn}\\
\texttt{\small\{198808xc, zxphistory\}@gmail.com} \; \texttt{\small tian.qi1@huawei.com}
}
\maketitle

\begin{abstract}
  Transformers have offered a new methodology of designing neural networks for visual recognition. Compared to convolutional networks, Transformers enjoy the ability of referring to global features at each stage, yet the attention module brings higher computational overhead that obstructs the application of Transformers to process high-resolution visual data. This paper aims to alleviate the conflict between efficiency and flexibility, for which we propose a specialized token for each region that serves as a messenger (\msg{}). Hence, by manipulating these \msg{} tokens, one can flexibly exchange visual information across regions and the computational complexity is reduced. We then integrate the \msg{} token into a multi-scale architecture named \name{}. In standard image classification and object detection, \name{} achieves competitive performance and the inference on both GPU and CPU is accelerated. Code is available at \url{https://github.com/hustvl/MSG-Transformer}.
\end{abstract}
{
\renewcommand{\thefootnote}{\fnsymbol{footnote}}
\footnotetext[2]{Corresponding author.}
\footnotetext[0]{The work was done during Jiemin Fang's internship at Huawei Inc.}}

\section{Introduction}
\label{sec:intro}

The past decade has witnessed the convolutional neural networks (CNNs) dominating the computer vision community. As one of the most popular models in deep learning, CNNs construct a hierarchical structure to learn visual features, and in each layer, local features are aggregated using convolutions to produce features of the next layer. Though simple and efficient, this mechanism obstructs the communication between features that are relatively distant from each other. To offer such an ability, researchers propose to replace convolutions by the Transformer, a module which is first introduced in the field of natural language processing~\cite{vaswani2017attention}. It is shown that Transformers have the potential to learn visual representations and achieve remarkable success in a wide range of visual recognition problems including image classification~\cite{ramachandran2019stand,dosovitskiy2020image}, object detection~\cite{carion2020end}, semantic segmentation~\cite{zheng2021setr}, \etc.

The Transformer module works by using a token to formulate the feature at each spatial position. The features are then fed into self-attention computation and each token, according to the vanilla design, can exchange information with all the others at every single layer. This design facilitates the visual information to exchange faster but also increases the computational complexity, as the computational complexity grows quadratically with the number of tokens -- in comparison, the complexity of a regular convolution grows linearly. To reduce the computational costs, researchers propose to compute attention in local windows of the 2D visual features. However constructing local attention within overlapped regions enables communications between different locations but causes inevitable memory waste and computation cost; computing attention within non-overlapped regions impedes information communications. As two typical local-attention vision Transformer methods, HaloNet~\cite{vaswani2021scaling} partitions query features without overlapping but overlaps key and value features by slightly increasing the window boundary; Swin Transformer~\cite{liu2021swin} builds implicit connections between windows by alternatively changing the partition style in different layers, \ie, shifting the split windows. These methods achieve competitive performance compared to vanilla Transformers, but HaloNet still wastes memories and introduces additional cost in the key and value; Swin Transformer relies on frequent 1D-2D feature transitions, which increase the implementation difficulty and additional latency.

To alleviate the burden, this paper presents a new methodology towards more efficient exchange of information. This is done by constructing a \textbf{messenger (\msg) token} in each local window. Each \msg\ token takes charge of summarizing information in the corresponding window and exchange it with other \msg\ tokens. In other words, all regular tokens are not explicitly connected to other regions, and \msg\ tokens serve as the hub of information exchange. This brings two-fold benefits. First, our design is friendly to implementation since it does not create redundant copies of data like \cite{ramachandran2019stand,zhao2020exploring,vaswani2021scaling}. Second and more importantly, the flexibility of design is largely improved. By simply manipulating the \msg\ tokens (\eg, adjusting the coverage of each messenger token or programming how they exchange information), one can easily construct many different architectures for various purposes. Integrating the Transformer with \msg{} tokens into a multi-scale design, we derive a powerful architecture named \textbf{\name{}} that takes advantages of both multi-level feature extraction and computational efficiency.

We instantiate \name{} as a straightforward case that the features of \msg\ tokens are shuffled and reconstructed with splits from different locations. This can effectively exchange information from local regions and delivered to each other in the next attention computation, while the implementation is easy yet efficient. We evaluate the models on both image classification and object detection, which achieve promising performance. We expect our efforts can further ease the research and application of multi-scale/local-attention Transformers for visual recognition.

We summarize our contributions as follows.
\begin{itemize}
\vspace{-5pt}
  \item We propose a new local-attention based vision Transformer with hierarchical resolutions, which computes attention in non-overlapped windows. Communications between windows are achieved via the proposed \msg\ tokens, which avoid frequent feature dimension transitions and maintain high concision and efficiency. The proposed shuffle operation effectively exchanges information from different \msg\ tokens with negligible cost.
\vspace{-5pt}
  \item In experiments, \name s show promising results on both ImageNet~\cite{imagenet} classification, \ie, 84.0\% Top-1 accuracy, and MS-COCO~\cite{COCO} object detection, \ie, 52.8 mAP, which consistently outperforms recent state-of-the-art Swin Transformer~\cite{liu2021swin}. Meanwhile, due to the concision for feature process, \name\ shows speed advantages over Swin Transformer, especially on the CPU device.
\vspace{-5pt}
  \item Not directly operating on the enormous patch tokens, we propose to use the lightweight \msg\ tokens to exchange information. The proposed \msg\ tokens effectively extract features from local regions and may have potential to take effects for other scenarios. We believe our work will be heuristic for future explorations on vision Transformers.
\end{itemize}

\section{Related Works}
\paragraph{Convolutional Neural Networks}
CNNs have been a popular and successful algorithm in a wide range of computer vision problems. As AlexNet~\cite{krizhevsky2012imagenet} shows strong performance on ImageNet~\cite{imagenet} classification, starting the blooming development of CNNs. A series of subsequent methods~\cite{szegedy2015going,simonyan2014very,he2016deep,szegedy2016rethinking,huang2017densely} emerge and persist in promoting CNN performance on vision tasks. Benefiting from the evolving of backbone networks, CNNs have largely improved the performance of various vision recognition scenarios including object detection~\cite{ren2015faster,redmon2016you,liu2016ssd,lin2017focal,cai2018cascade}, semantic/instance segmentation~\cite{DBLP:journals/pami/ChenPKMY18,deeplab-v3,he2017mask}, \etc. As real-life scenarios usually involve resource-constrained hardware platforms (\eg, for mobile and edge devices), CNNs are designed to take less computation cost~\cite{howard2017mobilenets,shufflev2,tan2019efficientnet}. Especially, with NAS approaches~\cite{zoph2018learning,cai2018proxylessnas,fbnet,fang2019densely} applied, CNNs achieved high performance with extremely low cost, \eg, parameter number, FLOPs and hardware latency. A clear drawback of CNNs is that it may take a number of layers for distant features to communicate with each other, hence limiting the ability of visual representation. Transformers aim to solve this issue.

\vspace{-8pt}
\paragraph{Vision Transformer Networks}
Transformers, first proposed by \cite{vaswani2017attention}, have been widely used in natural language processing (NLP). The variants of Transformers, together with improved frameworks and modules~\cite{devlin2018bert,gpt3}, have occupied most state-of-the-art (SOTA) performance in NLP. The core idea of Transformers lies in the self-attention mechanism, which aims at building relations between local features. Some preliminary works~\cite{wang2018non,huang2019ccnet,hu2019local,ramachandran2019stand,zhao2020exploring} explore to apply self-attention to networks for vision tasks and have achieved promising effects. Recently, ViT~\cite{dosovitskiy2020image} proposes to apply a pure Transformer to image patch sequences, which matches or even outperforms the concurrent CNN models on image classification. Inspired by ViT, a series of subsequent works~\cite{touvron2021training,yuan2021tokens,han2021transformer,chu2021we,touvron2021going} explore better designs of vision Transformers and achieve promising promotion. Some works~\cite{srinivas2021bottleneck,wu2021cvt,li2021bossnas,xu2021co} integrated modules from CNNs into vision Transformer networks and also achieve great results. In order to achieve strong results on image classification, many of the above ViT-based methods process features under a constant resolution and compute attentions within a global region. This makes it intractable to apply vision Transformers to downstream tasks, \eg, object detection and semantic segmentation, as multi-scale objects are hard to be represented under a constant resolution, and increased input resolutions cause overloaded computation/memory cost for attention computation.

\begin{figure*}[t!]
  \centering
  \includegraphics[width=\linewidth]{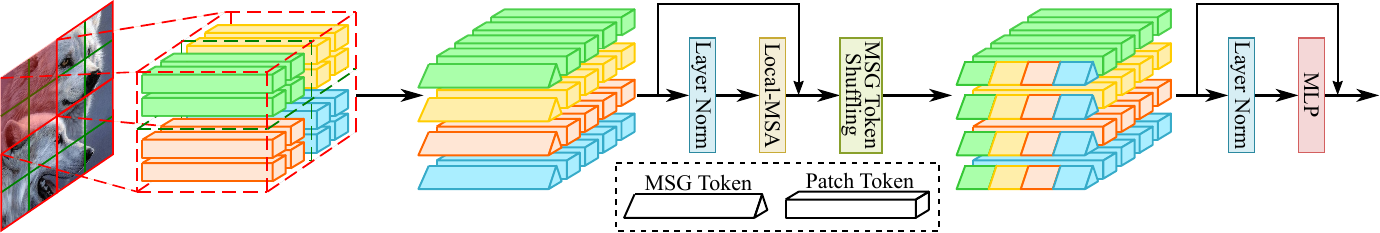}
  \caption{Structure of the \name\ block. The 2D features are split into local windows (by \textcolor{PineGreen}{green} lines), and several windows compose a shuffle region (the \textcolor{red}{red} one). Each local window is attached with one \msg\ token. \msg\ tokens are shuffled to exchange information in each Transformer block and deliver the obtained information to patch tokens in the next self-attention.}
  \label{fig: shuf_block}
  \vspace{-8pt}
\end{figure*}

\vspace{-8pt}
\paragraph{Downstream-friendly Vision Transformers}
To apply vision Transformers to downstream tasks, two key issues need to be solved, \ie, involving hierarchical resolutions to capture elaborate multi-scale features and decreasing cost brought by global attention computation. PVT~\cite{wang2021pyramid} proposed to process features under multi-resolution stages and down-samples key and value features to decrease the computation cost. HaloNet~\cite{vaswani2021scaling} and Swin Transformer~\cite{liu2021swin} propose to compute attention in a local window. To overcome the contradiction that non-overlapped windows lack communication while overlapped windows introduce additional memory/computation cost, HaloNet proposes to slightly overlap features in the key and value tokens but leave the query non-overlapped; Swin Transformer alternatively changes the window partition style to implicitly build connections between non-overlapped windows. A series of subsequent works~\cite{chu2021Twins,yu2021glance,huang2021shuffle,dong2021cswin} explore new methods for building local-global relations or connecting local regions.
We newly propose \msg\ tokens to extract information from local windows, and use a lightweight method, \ie, shuffle, to exchange information between \msg\ tokens. This concise manner avoids direct operation on cumbersome patch tokens and shows high flexibility.

\section{The \name{}}

This section elaborates the proposed approach, \name{}. The core part is Sec.~\ref{sec: shuf_block} where we introduce the \msg{} token and explain how it works to simplify information exchange. Then, we construct the overall architecture (\ie, the \name{}) in Sec.~\ref{sec: architecture} and analyze the complexity in Sec.~\ref{sec: complexity}.

\subsection{Adding MSG Tokens to a Transformer Block}
\label{sec: shuf_block}

The \name\ architecture is constructed by stacking a series of \name\ blocks, through various spatial resolutions. As shown in Fig.~\ref{fig: shuf_block}, a \name\ block mainly composes of several modules, \ie, layer normalization (layer norm), local multi-head self-attention (local-MSA), \msg\ token shuffling and MLP. 

Fig.~\ref{fig: shuf_block} presents how features from a local spatial region are processed. First, the 2D features $X \in \R^{H \times W \times C}$ are divided into non-overlapped windows (by \textcolor{PineGreen}{green} lines in Fig.~\ref{fig: shuf_block}) as $X_w \in \R^{\frac{H}{w} \times \frac{W}{w} \times w^2 \times C}$, where $(H, W)$ denotes the 2D resolution of the features, $C$ denotes the channel dimension, and $w$ denotes the window size. Then $R \times R$ windows compose a shuffle region (boxed in \textcolor{red}{red} lines in Fig.~\ref{fig: shuf_block}), namely features are split as $X_r \in \R^{\frac{H}{Rw} \times \frac{W}{Rw} \times R^2 \times w^2 \times C}$, where $R$ denotes the shuffle region. In vision Transformers~\cite{dosovitskiy2020image,touvron2021training}, image features are commonly projected into patch tokens by the input layer. Besides the patch tokens, which represent the intrinsic information of the images, we introduce an additional token, named messenger (\msg) token, to abstract information from patch tokens in a local window. Each local window is attached with one \msg\ token as $X_w' \in \R^{\frac{H}{Rw} \times \frac{W}{Rw} \times R^2 \times (w^2+1) \times C}$. Then a layer normalization is applied on all the tokens. The multi-head self-attention is performed within each local window between both patch and \msg\ tokens. \msg\ tokens can capture information from the corresponding windows with attention. Afterwards, all the \msg\ tokens $T_\mathrm{MSG} \in \R^{\frac{H}{Rw} \times \frac{W}{Rw} \times R^2 \times C}$ from a same local region $R \times R$ are shuffled to exchange information from different local windows. We name a region with \msg\ tokens shuffled as the \emph{shuffle region}. Finally, tokens are processed by a layer normalization and a two-layer MLP.

The whole computing procedure of a \name\ block can be summarized as follows.
\begin{align}
   X_w' &= [T_\mathrm{MSG}; X_w]\\
   X_w' &= \text{Local-MSA}(LN(X_w')) + X_w'\\
   T_\mathrm{MSG} &= \text{shuffle}(T_\mathrm{MSG})\\
   X_w' &= \text{MLP}(LN(X_w')) + X_w'
\end{align}

\begin{figure*}[thbp]
  \centering
  \includegraphics[width=0.9\linewidth]{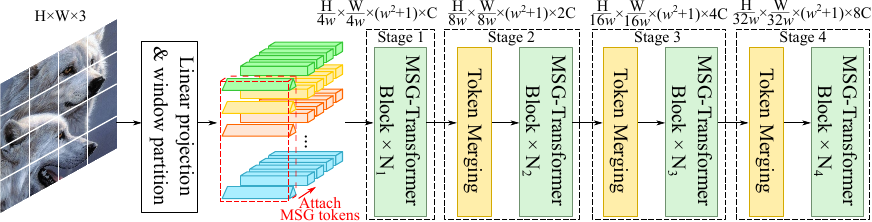}
  \caption{Overall architecture of \name. Patches from the input image are projected into tokens, and token features are partitioned into windows. Then each window is attached with one \msg\ token, which will participate in subsequent attention computation with all the other patch tokens within the local window in every layer. }
  \label{fig: arch}
\end{figure*}

\begin{figure}
\centering
\includegraphics[width=\linewidth]{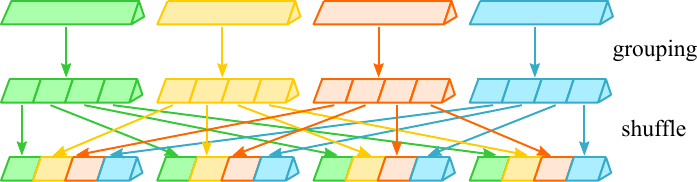}
\caption{Shuffling \msg\ tokens, where we inherit the example in Fig.~\ref{fig: shuf_block} for illustration.}
\vspace{-10pt}
\label{fig: shuf_op}
\end{figure}

\vspace{-8pt}
\paragraph{Local Multi-head Self-Attention}
Different from previous vision Transformers~\cite{dosovitskiy2020image,touvron2021training} which performer attention computation along the global region, we compute self-attention within each local window. Taking a window of $w \times w$ for example, the attention is computed on the token sequence of $X = [t_\mathrm{MSG}; x_1; ...; x_{w^2}]$, where $t_\mathrm{MSG}$ denotes the \msg\ token associated with this window and $x_i (1 \leq i \leq w^2)$ denotes each patch token within the window. 
\begin{equation}
    \text{Attention}(Q, K, V) = \text{softmax}(Q \cdot K^{T} / \sqrt{d} + B) \cdot V,
\end{equation}
where $Q, K, V \in \R^{(w^2+1)\times d}$ denotes the query, key and value matrices projected from sequence $X$ respectively, $d$ denotes the channel dimension, and $B$ denotes the relative position biases. Following previous Transformer works~\cite{hu2019local,raffel2020exploring,liu2021swin}, the relative position biases between patch tokens in $B$ are taken from the bias parameter $b^{rel} \in \R^{(2w-1)\times(2w-1)}$ according to the relative token distances. The position biases between patch tokens and the \msg\ token $t_\mathrm{MSG}$ is all set as equal, which is the same as the manner dealing with the \verb|[CLS]| token in \cite{ke2021rethinking}. Specifically, matrix $B$ are computed as
\begin{equation}
    B = \begin{cases}
        b^{rel}_{i', j'} & i \ne 0, j \ne 0\\
        \theta_1 & i=0 \\
        \theta_2 & i \ne 0, j=0
    \end{cases},
\end{equation}
where $i' = i\operatorname{mod}w - j\operatorname{mod}w + w - 1, j' = i//w - j//w + w - 1$, $\theta_1, \theta_2$ are two learnable parameters.

\vspace{-8pt}
\paragraph{Exchanging Information by Shuffling \msg\ Tokens}
The \msg{} tokens allow us to exchange visual information flexibly. Here we instantiate an example using the shuffling operation, while we emphasize that the framework easily applies to other operations (see the next paragraph). In each \name\ block, \msg\ tokens in a same shuffle region are shuffled to exchange information from different local windows. Assuming the shuffle region has a size of $R \times R$, it means there exist $R \times R$ \msg\ tokens in this region and each \msg\ token is associated with a $w \times w$ local window. As shown in Fig~\ref{fig: shuf_op}, channels of each \msg\ token are first split into $R \times R$ groups. Then the groups from $R \times R$ \msg\ tokens are recombined. With shuffle finished, each \msg\ token obtains information from all the other ones. With the next attention computing performed, spatial information from the other local windows is delivered to patch tokens in the current window via the \msg\ token. Denoting \msg\ tokens in a $R \times R$ shuffle region as $T_\mathrm{MSG} \in \R^{R^2 \times d}$, the shuffle process can be formulated as
\begin{equation}
\begin{aligned}
    T_\mathrm{MSG}' &= \text{reshape}(T_\mathrm{MSG}), \; T_\mathrm{MSG}' \in \R^{R^2 \times R^2 \times \frac{d}{R^2}}\\
    T_\mathrm{MSG}' &= \text{transpose}(T_\mathrm{MSG}', \text{dim}_0=0, \text{dim}_1=1)\\
    T_\mathrm{MSG} &= \text{reshape}(T_\mathrm{MSG}'), \; T_\mathrm{MSG} \in \R^{R^2 \times d}
\end{aligned},
\end{equation}
where $d$ denotes the channel dimension of the \msg\ token, which is guaranteed to be divisible by the group number, $R^2$.

Though the shuffle operation has the similar manner with that in convolutional network ShuffleNet~\cite{zhang2017shufflenet,shufflev2}, the effect is entirely different. ShuffleNet performs the shuffle operation to fuse separated channel information caused by the grouped $1 \times 1$ convolution, while our \name\ shuffles the proposed \msg\ tokens to exchange spatial information from different local windows. 

\vspace{-8pt}
\paragraph{Extensions}
There exist other ways of constructing and manipulating \msg{} tokens. For example, one can extend the framework so that neighboring \msg{} tokens can overlap, or program the propagation rule so that the \msg{} tokens are not fully-connected to each other. Besides, one can freely inject complex operators, rather than shuffle-based identity mapping, when the features of \msg{} tokens are exchanged. Note that some of these functions are difficult to implement without taking \msg{} tokens as the explicit hub. We will investigate these extensions in the future.

\subsection{Overall Architecture}
\label{sec: architecture}

\begin{table*}[h]
  \centering
  \caption{Detailed settings for \name\ architecture variants. `cls' denotes image classification on ImageNet. `det' denotes object detection on MS-COCO. `dim' denotes the embedding channel dimension. `\#head' and `\#blocks' denote the number of self-attention heads and \name\ blocks in each stage.}
  \small
  \begin{tabular}{c|c|c|c|c|c|c|c|c|c|c|c|c}
  \toprule
  \multirow{2}*{\textbf{Stage}} & \textbf{Patch Token} & \multicolumn{2}{c|}{\textbf{Shuffle Size}} & \multicolumn{3}{c|}{\textbf{\name{}-T}} & \multicolumn{3}{c|}{\textbf{\name{}-S}} & \multicolumn{3}{c}{\textbf{\name{}-B}}\\
  \cline{3-13}
  & \textbf{Resolution} & cls & det & dim & \#heads & \#blocks & dim & \#heads & \#blocks &  dim & \#heads & \#blocks \\
  \hline
  1 & \begin{tabular}{c} \spaceline $\frac{H}{4} \times \frac{W}{4}$ \\ \spaceline\end{tabular} & 4 & 4 & 64 & 2 & 2 & 96 & 3 & 2 & 96 & 3 & 2 \\ 
  \hline
  2 & \begin{tabular}{c} \spaceline $\frac{H}{8} \times \frac{W}{8}$ \\ \spaceline\end{tabular} & 4 & 4 & 128 & 4 & 4 & 192 & 6 & 4 & 192 & 6 & 4 \\ 
  \hline
  3 & \begin{tabular}{c} \spaceline $\frac{H}{16} \times \frac{W}{16}$ \\ \spaceline\end{tabular} & 2 & 8 & 256 & 8 & 12 & 384 & 12 & 12 & 384 & 12 & 28 \\ 
  \hline
  4 & \begin{tabular}{c} \spaceline $\frac{H}{32} \times \frac{W}{32}$ \\ \spaceline\end{tabular} & 1 & 4 & 512 & 16 & 4 & 768 & 24 & 4 & 768 & 24 & 4 \\ 
  \bottomrule
  \end{tabular}
  \label{tab: arch_detail}
\end{table*}

Fig.~\ref{fig: arch} shows the overall architecture of \name. The input image is first projected into patch tokens $T_p \in \R^{\frac{H}{4} \times \frac{W}{4} \times C}$ by a $7 \times 7$ convolution with stride $4$, where $C$ denotes the channel dimension. The overlapped projection is used for building better relations between patch tokens. Similar manners have also been adopted in previous methods~\cite{wu2021cvt,chu2021Twins}. Then the tokens are split into windows with the shape of $w \times w$, and each window is attached with one \msg\ token, which has an equal channel number with the patch token. The rest part of the architecture is constructed by stacking a series of \name\ blocks as defined in Sec.~\ref{sec: shuf_block}. To obtain features under various spatial resolutions, we downsample features by merging both patch and \msg\ tokens. Blocks under the same resolution form a stage. For both patch and \msg\ tokens, we use an overlapped $3 \times 3$ convolution with stride $2$ to perform token merging and double the channel dimension in the next stage\footnote{The convolution parameters for merging tokens are shared between patch and \msg\ tokens.}. For image classification, the finally merged \msg\ tokens are projected to produce classification scores. And for downstream tasks like object detection, only patch tokens are delivered into the head structure while \msg\ tokens only serve for exchanging information in the backbone.

In our implementation, we build three architecture variants with different scales. As shown in Tab.~\ref{tab: arch_detail}, \name-T, -S and -B represent tiny, small, and base architectures with different channel numbers, attention head numbers and layer numbers.  The window size is set as $7$ for all architectures. The shuffle region size is set as $4, 4, 2, 1$ in four stages respectively for image classification and $4, 4, 8, 4$ for object detection. As demonstrated in subsequent studies (Sec.~\ref{sec: exp_abl}), our \name{} prefers deeper and narrower architecture scales than Swin Transformer~\cite{liu2021swin}.

\subsection{Complexity Analysis}
\label{sec: complexity}

Though introduced one \msg\ token in each local window, the increased computational complexity is negligible. The local attention-based Transformer block includes two main part, \ie, local-MSA and two-layer MLP. Denoting the input patch token features as $T_p \in \R^{\frac{H}{w}\times\frac{W}{w}\times w^2 \times C}$, where $H, W$ denote the 2D spatial resolution, $w$ denotes the local window size, and $C$ denotes the channel number, the total FLOPs are computed as 
\begin{equation}
\begin{aligned}
  &\text{FLOPs} = \text{FLOPs}_{MSA} + \text{FLOPs}_{MLP}\\
  &= \frac{HW}{w^2} \times (4w^2C^2 + 2w^4C) +2 \frac{HW}{w^2}w^2 \cdot 4C^2.
\end{aligned}
\end{equation}
With the \msg\ tokens applied, the total FLOPs change to
\begin{equation}
\begin{aligned}
  \text{FLOPs}' =& \frac{HW}{w^2}(4(w^2+1)C^2 + 2(w^2+1)^2C) \\
  &+ 2 \frac{HW}{w^2}(w^2+1) \cdot 4C^2.
\end{aligned}
\end{equation}
The FLOPs increase proportion is computed as
\begin{equation}
\begin{aligned}
  &\frac{\text{FLOPs}' - \text{FLOPs}}{\text{FLOPs}} \\
  &= \frac{\frac{HW}{w^2}(4C^2 + 2(w^2+1)C) + 2\frac{HW}{w^2} \cdot 4C^2}
  {\frac{HW}{w^2} \times (4w^2C^2 + 2w^4C) +2 \frac{HW}{w^2} \times w^2 \times 4C^2} \\
  &= \frac{6C+w^2+1}{6w^2C+w^4}.
\end{aligned}
\end{equation}
As the window size $w$ is set as $7$ in our implementations, the FLOPs increase proportion becomes $\frac{6C+50}{294C+7^4}$. Taking the channel number as $384$ for example, the increased FLOPs only account for $\sim2.04\%$ which are negligible to the total complexity. 

For the number of parameters, all the linear projection parameters are shared between patch and \msg\ tokens. Only the input \msg\ tokens introduce additional parameters, but they are shared between shuffle regions, only taking $4^2C = 16C$, \ie, $\sim0.0015$M for the $96$ input channel dimension. In experiments, we prove even with the input \msg\ tokens not learned, \name s can still achieve as high performance. From this, parameters from input \msg\ tokens can be abandoned.

It is worth noting that due to local region communication is achieved by shuffling \msg\ tokens, the huge feature matrix of patch tokens only needs to be window-partitioned once in a stage if the input images have a regular size. With \msg\ tokens assisting, cost from frequent 2D-to-1D matrix transitions of patch tokens can be saved, which cause additional latencies especially on computation-limited devices, \eg, CPU and mobile devices, but are unavoidable in most previous local attention-based~\cite{ramachandran2019stand,liu2021swin} or CNN-attention hybrid Transformers~\cite{chu2021we,wu2021cvt,li2021bossnas}.

\begin{table}[t]
\centering
\small
\caption{Image classification performance comparisons on ImageNet-1K~\cite{imagenet}.}
\label{tab: imagenet}
\resizebox{\columnwidth}{!}{
\begin{threeparttable}
\setlength{\tabcolsep}{0.1cm}
\begin{tabular}{l|c|c|c|c|c|c}
\toprule
\textbf{Method} &  \begin{tabular}[c]{@{}c@{}}\textbf{Input} \\ \textbf{size}\end{tabular} & \textbf{Params} & \textbf{FLOPs} & \textbf{Imgs/s} & \begin{tabular}[c]{@{}c@{}}\textbf{CPU}\\ \textbf{latency}\end{tabular} & \begin{tabular}[c]{@{}c@{}}\textbf{Top-1}\\ \textbf{(\%)}\end{tabular} \\
\midrule
\multicolumn{7}{l}{\textbf{Convolutional Networks}}\\
RegY-4G~\cite{radosavovic2020designing} & 224$^2$ & 21M & 4.0G & 930.1 & 138ms & 80.0 \\
RegY-8G~\cite{radosavovic2020designing} & 224$^2$ & 39M & 8.0G & 545.5 & 250ms & 81.7 \\
RegY-16G~\cite{radosavovic2020designing} & 224$^2$ & 84M & 16.0G & 324.6 & 424ms & 82.9 \\
EffNet-B4~\cite{tan2019efficientnet} & 380$^2$ & 19M & 4.2G & 345 & 315ms & 82.9 \\
EffNet-B5~\cite{tan2019efficientnet} & 456$^2$ & 30M & 9.9G & 168.5 & 768ms & 83.6 \\
EffNet-B6~\cite{tan2019efficientnet} & 528$^2$ & 43M & 19.0G & 96.4 & 1317ms & 84.0 \\
\midrule
\multicolumn{7}{l}{\textbf{Transformer Networks}}\\
DeiT-S~\cite{touvron2021training} & 224$^2$ & 22M & 4.6G & 898.3 & 118ms & 79.8 \\
T2T-ViT$_t$-14~\cite{yuan2021tokens} & 224$^2$ & 22M & 5.2G & 559.3 & 225ms & 80.7\\
PVT-Small~\cite{wang2021pyramid} & 224$^2$ & 25M & 3.8G & 749.0 & 146ms & 79.8 \\
TNT-S~\cite{han2021transformer} & 224$^2$ & 24M & 5.2G & 387.1 & 215ms & 81.3\\
CoaT-Lite-S~\cite{xu2021co} & 224$^2$ & 20M & 4.0G & - & - & 81.9 \\
Swin-T~\cite{liu2021swin} & 224$^2$ & 28M & 4.5G & 692.1 & 189ms & 81.3 \\
MSG-T & 224$^2$ & 25M & 3.8G & 726.5 & 157ms & \textbf{82.4} \\
\midrule
DeiT-B~\cite{touvron2021training} & 224$^2$ & 87M & 17.5G & 278.9 & 393ms & 81.8 \\
T2T-ViT$_t$-19~\cite{yuan2021tokens} & 224$^2$ & 39M & 8.4G & 377.3 & 314ms & 81.4\\
T2T-ViT$_t$-24~\cite{yuan2021tokens} & 224$^2$ & 64M & 13.2G & 268.2 & 436ms & 82.2\\
PVT-Large~\cite{wang2021pyramid} & 224$^2$ & 61M & 9.8G & 337.1 & 338ms & 81.7 \\
TNT-B~\cite{han2021transformer} & 224$^2$ & 66M & 14.1G & 231.1 & 414ms & 82.8\\
Swin-S~\cite{liu2021swin} & 224$^2$ & 50M & 8.7G & 396.6 & 346ms & 83.0 \\
MSG-S & 224$^2$ & 56M & 8.4G & 422.5 & 272ms & \textbf{83.4} \\
\midrule
ViT-B/16~\cite{dosovitskiy2020image} & 384$^2$ & 87M & 55.4G & 81.1 & 1218ms & 77.9 \\
ViT-L/16~\cite{dosovitskiy2020image} & 384$^2$ & 307M & 190.7G & 26.3 & 4420ms & 76.5 \\
DeiT-B~\cite{touvron2021training} & 384$^2$ & 87M & 55.4G & 81.1 & 1213ms & 83.1 \\
Swin-B~\cite{liu2021swin} & 224$^2$ & 88M & 15.4G & 257.6 & 547ms & 83.3 \\
MSG-B & 224$^2$ & 84M & 14.2G & 267.6 & 424ms & \textbf{84.0} \\
\bottomrule
\end{tabular}
\begin{tablenotes}
  \item[*] ``Imgs/s'' denotes the GPU throughput which is measured on one 32G-V100 with a batch size of 64. Noting that throughput on 32G-V100 used in our experiments is sightly lower than 16G-V100 used in some other papers.
  \item[*] The CPU latency is measured with one core of Intel(R) Xeon(R) Gold 6151 CPU @ 3.00GHz.
\end{tablenotes}
\end{threeparttable}
}
\end{table}
\vspace{-5pt}

\section{Experiments}
In experiments, we first evaluate our \name\ models on ImageNet~\cite{imagenet} classification in Sec.~\ref{sec: exp_cls}. Then in Sec.~\ref{sec: exp_obj}, we evaluate \name s on MS-COCO~\cite{COCO} object detection and instance segmentation. Finally, we perform a series of ablation studies and analysis in Sec.~\ref{sec: exp_abl}. Besides, we provide a MindSpore \cite{mindspore} implementation of \name.

\subsection{Image Classification}
\label{sec: exp_cls}
We evaluate our \name\ networks on the commonly used image classification dataset ImageNet-1K~\cite{imagenet} and report the accuracies on the validation set in Tab.~\ref{tab: imagenet}. Most training settings follow DeiT~\cite{touvron2021training}. The AdamW~\cite{DBLP:journals/corr/KingmaB14} optimizer is used with $0.05$ weight decay. The training process takes $300$ epochs in total with a cosine annealing decay learning rate schedule~\cite{DBLP:conf/iclr/LoshchilovH17} and 20-epoch linear warmup. The total batch size is set as $1024$ and the initial learning rate is $0.001$. The repeated augmentation~\cite{hoffer2020augment} and EMA~\cite{polyak1992acceleration} are not used as in Swin Transformer~\cite{liu2021swin}.

\begin{table}
\caption{Object detection and instance segmentation performance comparisons on MS-COCO~\cite{COCO} with Cascade Mask R-CNN~\cite{he2017mask,cai2018cascade}. ``X101-32'' and ``X101-64'' denote ResNeXt101-32$\times$4d~\cite{xie2017aggregated} and -64$\times$4d respectively.}
\centering
\small
\label{tab: coco}
\resizebox{\columnwidth}{!}{
\begin{threeparttable}
\setlength{\tabcolsep}{0.02cm}
\begin{tabular}{l|ccc|ccc|ccc}
  \toprule
  \textbf{Method} & \textbf{AP$^\text{box}$} & \textbf{AP$^\text{box}_\text{50}$} & \textbf{AP$^\text{box}_\text{75}$} & \textbf{AP$^\text{mask}$} & \textbf{AP$^\text{mask}_\text{50}$} & \textbf{AP$^\text{mask}_\text{75}$} & \textbf{Params} & \textbf{FLOPs} & \textbf{FPS} \\
  \midrule
  DeiT-S & 48.0 & 67.2 & 51.7 & 41.4 & 64.2 & 44.3 & 80M & 889G & - \\
  ResNet-50 & 46.3 & 64.3 & 50.5 & 40.1 & 61.7 & 43.4 & 82M & 739G & 10.5 \\
  Swin-T & 50.5 & 69.3 & 54.9 & 43.7 & 66.6 & 47.1 & 86M & 745G & 9.4 \\
  MSG-T & \textbf{51.4} & \textbf{70.1} & \textbf{56.0} & \textbf{44.6} & \textbf{67.4} & \textbf{48.1} & 83M & 731G & 9.1 \\
  \midrule
  X101-32 & 48.1 & 66.5 & 52.4 & 41.6 & 63.9 & 45.2 & 101M & 819G & 7.5 \\
  Swin-S & 51.8 & 70.4 & 56.3 & 44.7 & 67.9 & 48.5 & 107M & 838G & 7.5 \\
  MSG-S & \textbf{52.5} & \textbf{71.1} & \textbf{57.2} & \textbf{45.5} & \textbf{68.4} & \textbf{49.5} & 113M & 831G & 7.5 \\
  \midrule
  X101-64 & 48.3 & 66.4 & 52.3 & 41.7 & 64.0 & 45.1 & 140M & 972G & 6.0 \\
  Swin-B & 51.9 & 70.9 & 56.5 & 45.0 & 68.4 & 48.7 & 145M & 982G & 6.3 \\
  MSG-B & \textbf{52.8} & \textbf{71.3} & \textbf{57.3} & \textbf{45.7} & \textbf{68.9} & \textbf{49.9} & 142M & 956G & 6.1 \\
  \bottomrule
\end{tabular}
\begin{tablenotes}
    \item[*] FPS is measured on one 32G-V100 with a batch size of $1$.
\end{tablenotes}
\end{threeparttable}
}
\vspace{-10pt}
\end{table}

We provide the ImageNet classification results in Tab.~\ref{tab: imagenet} and compare with other convolutional and Transformer networks. Compared with DeiT~\cite{liu2019training}, \name s achieve significantly better trade-offs between accuracy and computation budget. \name-T achieves $2.6$ Top-1 accuracy promotion over DeiT-S with $0.8$G smaller FLOPs; \name-S promotes the accuracy by $1.6$ with only $48.0\%$ FLOPs; \name-B achieves an $84.0\%$ Top-1 accuracy, beating larger-resolution DeiT-B by $0.9$ with only $25.6\%$ FLOPs. Compared with the recent state-of-the-art method Swin Transformer~\cite{liu2021swin}, our \name s achieve competitive accuracies with similar Params and FLOPs. It is worth noting, as frequent 1D-2D feature transitions and partition are avoided, \name s show promising speed advantages over Swin Transformers. Especially on the CPU device, the latency improvement is more evident. \name-T is $16.9\%$ faster than Swin-T; \name-S is $21.4\%$ faster than Swin-S; \name-B is $22.5\%$ faster than Swin-B.

\subsection{Object Detection}
\label{sec: exp_obj}
We evaluate our \name\ networks on MS-COCO~\cite{COCO} object detection with the Cascade Mask R-CNN~\cite{cai2018cascade,he2017mask} framework. The training and evaluation are performed based on the MMDetection~\cite{chen2019mmdetection} toolkit. For training, we use the AdamW~\cite{DBLP:journals/corr/KingmaB14} optimizer with $0.05$ weight decay, $1\times10^{-4}$ initial learning rate and a total batch size of 16. The learning rate is decayed by $0.1$ at the $27$ and $33$ epoch. The training takes the $3\times$ schedule, \ie, 36 epochs in total. Multi-scale training with the shorter side of the image resized between $480$ and $800$ and the longer side not exceeding $1333$ is also used. As the input image size is not fixed for object detection, the patch tokens are padded with $0$ to guarantee they can be partitioned by the given window size for attention computation. And the shuffle region is alternatively changed at the left-top and right-bottom locations between layers to cover more windows. 

As shown in Tab.~\ref{tab: coco}, \name s achieve significantly better performance than CNN-based models, \ie, $5.1$ AP$^\text{box}$ better than ResNet-50~\cite{he2016deep}, $4.4$ AP$^\text{box}$ better than ResNeXt101-32$\times$4d~\cite{xie2017aggregated}, and $4.5$ AP$^\text{box}$ better than ResNeXt101-64$\times$4d. Even though Swin Transformers have achieved extremely high performance on object detection, our \name s still achieve significant promotion by $0.9$, $0.7$, $0.9$ AP$^\text{box}$ and $0.9$, $0.8$, $0.7$ AP$^\text{mask}$ for T, S, B scales respectively. 

\subsection{Ablation Study}
\label{sec: exp_abl}
In this section, we perform a series of ablation studies on ImageNet-1K about the shuffling operation, \msg\ tokens, network scales, and shuffle region sizes\footnote{Without specified, experiments for ablation study remove the overlapped downsampling and follow the network scales in Swin-Transformer~\cite{liu2021swin} for clear and fair comparisons.}. We further visualize the attention map of \msg\ tokens for better understanding the working mechanism.

\begin{table}
\centering
\small
\caption{Ablation studies about \msg\ tokens and shuffle operations on ImageNet classification.}
\label{tab: shuf_msg_effect}
\resizebox{\columnwidth}{!}{
\begin{tabular}{c|c|c|c|l}
  \toprule
  \textbf{Row} & \textbf{\msg\ Token} & \textbf{Shuffle Op.} & \textbf{Images / s} & \textbf{Top1 (\%)} \\
  \midrule
  \multicolumn{5}{l}{\textbf{\name{}-T (depth=12)}}\\
  1 & \xmark & \xmark & 720.3 & 80.2 \\
  2 & \checkmark & \xmark & 702.2 & 80.5$_{\uparrow 0.3}$ \\
  3 & \checkmark & \checkmark & 696.7 & \textbf{81.1}$_{\uparrow 0.9}$ \\
  \midrule
  \multicolumn{5}{l}{\textbf{\name{}-S (depth=24)}}\\
  4 & \xmark & \xmark & 412.9 & 81.2 \\
  5 & \checkmark & \xmark & 403.9 & 81.9$_{\uparrow 0.7}$ \\
  6 & \checkmark & \checkmark & 401.0 & \textbf{83.0}$_{\uparrow 1.8}$ \\
  \bottomrule
\end{tabular}}
\vspace{-10pt}
\end{table}

\vspace{-8pt}
\paragraph{Effects of \msg\ Tokens \& Shuffle Operations}
We study the effects of \msg\ tokens and shuffle operations, providing the results in Tab.~\ref{tab: shuf_msg_effect}. As shown in Row 1, with both MSG tokens and shuffle operations removed, the performance degrades by $0.9$. With \msg\ tokens applied in Row 2, the performance is promoted by $0.3$ compared to that without both. Though without shuffle operations, \msg\ tokens can still exchange information in each token merging (downsampling) layer, which leads to slight promotion. However, exchanging information only in token merging layers is too limited to expanding receptive fields. With the same ablation applied on a deeper network \name-S, the performance gap becomes significant. The Top-1 accuracy drops by $1.8$ with both modules removed, and drops by $1.1$ with shuffle removed. It is worth noting that both \msg\ tokens and shuffle operations are light enough and cause no evident throughput decay. 

\begin{table}
\caption{Effects of input \msg/\texttt{CLS} token parameters on ImageNet classification.}
\centering
\small
\label{tab: msg_learn}
\begin{tabular}{l|c|c|l}
  \toprule
  \textbf{Row} & \textbf{Training} & \textbf{Evaluation} & \textbf{Top1 (\%)} \\
  \midrule
  \multicolumn{4}{l}{\textbf{\name-T (\msg\ Token)}}\\
  1 & learnable & learned & 80.9 \\
  2 & learnable & random & 80.8$_{\downarrow 0.1}$ \\
  3 & random & random & 80.8$_{\downarrow 0.1}$ \\
  \midrule
  \multicolumn{4}{l}{\textbf{Deit-S (\texttt{CLS} Token)}}\\
  4 & learnable & learned & 79.9 \\
  5 & learnable & random & 77.7$_{\downarrow 2.2}$ \\
  \bottomrule
\end{tabular}
\vspace{-15pt}
\end{table}

\begin{figure*}[htbp]
\vspace{-5pt}
\centering
\begin{minipage}[b]{0.15\linewidth}
\includegraphics[width=1\linewidth]{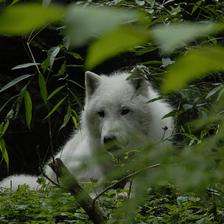}
\centerline{\small Input image}
\end{minipage}\hfill
\begin{minipage}[b]{0.15\linewidth}
\includegraphics[width=1\linewidth]{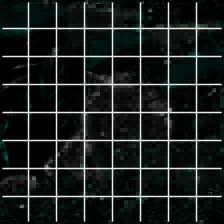}
\centerline{\small Block-2}
\end{minipage}\hfill
\begin{minipage}[b]{0.15\linewidth}
\includegraphics[width=1\linewidth]{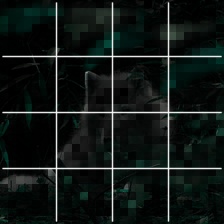}
\centerline{\small Block-4}
\end{minipage}\hfill
\begin{minipage}[b]{0.15\linewidth}
\includegraphics[width=1\linewidth]{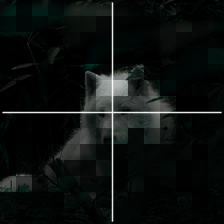}
\centerline{\small Block-7}
\end{minipage}\hfill
\begin{minipage}[b]{0.15\linewidth}
\includegraphics[width=1\linewidth]{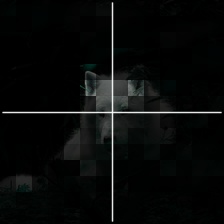}
\centerline{\small Block-10}
\end{minipage}\hfill
\begin{minipage}[b]{0.15\linewidth}
\includegraphics[width=1\linewidth]{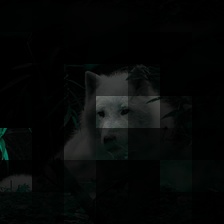}
\centerline{\small Block-12}
\end{minipage}
\vspace{-3pt}
\caption{Visualization of attention maps computed between each \msg{} token and patch tokens within the local window in different blocks.}
\label{fig: vis}
\vspace{-8pt}
\end{figure*}

\vspace{-8pt}
\paragraph{Input Parameters of \msg\ Tokens}
To further understand the role \msg\ tokens play in Transformers, we study impacts caused by parameters of \textbf{input} \msg\ tokens. As shown in Row 2 of Tab.~\ref{tab: msg_learn}, we randomly re-initialize parameters of input \msg\ tokens for evaluation, which are learnable during training, interestingly the accuracy only drops by $0.1\%$. Then in Row 3, we randomly initialize input \msg\ tokens and keep them fixed during training. It still induces negligible accuracy drop when input \msg\ token parameters are also randomly re-initialized for evaluation. This implies input \msg\ token parameters are not so necessary to be learned. We speculate that if input parameters of \verb|CLS| tokens in conventional Transformers need to be learned, and perform the same experiment on Deit-S~\cite{touvron2021training}. Then we find randomly re-parameterizing input \verb|CLS| tokens for evaluation leads to severe degradation to the accuracy, \ie, 2.2\% in Row 5. 

The above experiments show that the proposed \msg\ tokens play a different role from conventional \verb|CLS| tokens, which serve as messengers to carry information from different local windows and exchange information with each other. The input parameters of their own matter little in latter information delivering as they absorb local features layer by layer via attention computing. In other words, with uninterrupted self-attention and information exchanging, patch tokens make what \msg\ tokens are and \msg\ tokens are just responsible for summarizing local patch tokens and deliver the message to other locations. Therefore, input parameters of \msg\ tokens do not affect the final performance.

\vspace{-10pt}
\paragraph{Network Scales}

Considering different types of architectures fit different network scales, we study the scales of both Swin- and MSG-Transformer as follows. As shown in Tab.~\ref{tab: net_scale}, two scales are evaluated where one is shallow and wide with $96$ input dimension and $[2, 2, 6, 2]$ blocks in each stage, while another one is deep and narrow with $64$ dimension and $[2, 4, 12, 4]$ blocks. We observe \name{} achieves a far better trade-off between computation cost and accuracy with the deeper-narrower scale. We analyze the reason as follows. In Swin Transformer, each patch token is involved in two different windows between layers, which requires a wider channel dimension with more attention heads to support the variety. On the contrary, \name{} uses MSG tokens to extract window-level information and transmit to patch tokens. This reduces the difficulty for patch tokens to extract information from other windows. Thus \name\ requires a smaller channel capacity to support variety in one window. A deeper and narrower architecture brings a better trade-off for \name.

\begin{table}
\centering
\small
\caption{Network scale study on Swin- and MSG-Transformer.}
\label{tab: net_scale}
\resizebox{\columnwidth}{!}{
\begin{tabular}{c|c|cccc}
  \toprule
  \textbf{Model} & \textbf{Dim} & \textbf{\#Blocks} & \textbf{Params} & \textbf{FLOPs} & \textbf{Top1 (\%)} \\
  \midrule
  \multirow{2}{*}{Swin} & 96 & [2, 2, 6, 2] & 28M & 4.5G & 81.3 \\
   & 64 & [2, 4, 12, 4] & 24M & 3.6G & 81.3 \\
  \midrule
  \multirow{2}{*}{MSG} & 96 & [2, 2, 6, 2] & 28M & 4.6G & 81.1 \\
   & 64 & [2, 4, 12, 4] & 24M & 3.7G & \textbf{82.1} \\
  \bottomrule
\end{tabular}
}

\end{table}
\begin{table}
\centering
\caption{Ablation studies about the shuffle region sizes on ImageNet classification.}
\label{tab: shuf_size}
\small
\begin{tabular}{c|c|cccc}
  \toprule
  \textbf{Shuffle Region Sizes} & \textbf{Images / s} & \textbf{Top1(\%)} \\
  \midrule
  $2, 2, 2, 1$ & 695.1 & 80.6 \\
  $4, 2, 2, 1$ & 696.1 & 80.8 \\
  $4, 4, 2, 1$ & 696.7 & \textbf{80.9} \\
  \bottomrule
\end{tabular}
\vspace{-10pt}
\end{table}

\vspace{-10pt}
\paragraph{Shuffle Region Sizes}
We study the impacts of shuffle region sizes on the final performance. As shown in Tab.~\ref{tab: shuf_size}, with the shuffle region enlarged, the final accuracy increases. It is reasonable that larger shuffle region sizes lead to larger receptive fields and are beneficial for tokens capturing substantial spatial information. Moreover, the throughput/latency is not affected by the shuffle size changing.

\vspace{-10pt}
\paragraph{Attention Map Visualization of \msg\ Tokens}
To understand the working mechanism of \msg\ tokens, we visualize attention maps computed between each \msg\ token and its associated patch tokens within the local window in different blocks. As shown in Fig.~\ref{fig: vis}, local windows in attention maps are split into grids. Though the local window size to the token features is constant, \ie $7$ in our settings, with tokens merged, the real receptive field is enlarged when reflected onto the original image. In shallower blocks, attention of \msg\ tokens is dispersive which tends to capture contour information; in deeper layers, though attention is computed within each local window, \msg{} tokens can still focus on locations closely related to the object.

\section{Discussion and Conclusion}
This paper proposes \name{}, a novel Transformer architecture that enables efficient and flexible information exchange. The core innovation is to introduce the \msg{} token which serves as the hub of collecting and propagating information. We instantiate \name{} by shuffling \msg{} tokens, yet the framework is freely extended by simply altering the way of manipulating \msg{} tokens. Our approach achieves competitive performance on standard image classification and object detection tasks with reduced implementation difficulty and faster inference speed.

\vspace{-10pt}
\paragraph{Limitations} We would analyze limitations from the perspective of the manipulation type for MSG tokens. Though shuffling is an efficient communication operation, the specificity of shuffled tokens is not so well as shuffling integrates token segments from different local windows equally on the channel dimension. On the other hand, it is valuable to explore other manipulation types with a better efficiency-specificity trade-off which may further motivate the potential of \name.

\vspace{-10pt}
\paragraph{Future work} Our design puts forward an open problem: since information exchange is the common requirement of deep networks, how to satisfy all of capacity, flexibility, and efficiency in the architecture design? The \msg{} token offers a preliminary solution, yet we look forward to validating its performance and further improving it in visual recognition tasks and beyond.

\section*{Acknowledgement}
We thank Yuxin Fang, Bencheng Liao, Liangchen Song, Yuzhu Sun and Yingqing Rao for constructive discussions and assistance. This work was in part supported by NSFC (No. 61876212 and No. 61733007) and CAAI-Huawei MindSpore Open Fund.

{\small
\bibliographystyle{ieee_fullname}
\bibliography{egbib}
}

\appendix
\section{Appendix}

\begin{table}[t!]
\centering
\small
\caption{Image classification accuracy on ImageNet-1K comparing with concurrent hierarchical Transformers.}
\label{tab: imagenet_con}
\begin{threeparttable}
\setlength{\tabcolsep}{4pt}
\begin{tabular}{c|l|c|c|c}
\toprule
\textbf{w/ dw-conv?} & \textbf{Method} &  \textbf{Params} & \textbf{FLOPs} & \textbf{Top-1 (\%)} \\
\midrule
\multirow{2}{*}{\xmark} & Swin-T~\cite{liu2021swin} & 28M & 4.5G & 81.3 \\
& MSG-T & 25M & 3.8G & \textbf{82.4} \\
\midrule
\multirow{4}{*}{\checkmark} & GG-T~\cite{yu2021glance} & 28M & 4.5G & 82.0 \\
& Shuffle-T~\cite{huang2021shuffle} & 29M & 4.6G & 82.5 \\
& CSWin-T~\cite{dong2021cswin} & 23M & 4.3G & 82.7 \\
& MSG-T$_\mathrm{dwc}$ & 25M & 3.9G & \textbf{83.0} \\
\bottomrule
\end{tabular}
\end{threeparttable}
\end{table}
\subsection{Comparisons with Concurrent Hierarchical Transformers}
Some related concurrent works~\cite{yu2021glance,huang2021shuffle,dong2021cswin} also focus on improving attention computing patterns with different manners based on a hierarchical architecture and achieve remarkable performance. These works introduce additional depth-wise convolutions~\cite{chollet2017xception} into Transformer blocks, which improve recognition accuracy with low FLOPs increase. Our \name{}s in the main text do not include depth-wise convolutions to make the designed model a purer Transformer. We further equip MSG-T with depth-wise convolutions, resulting in a variant named MSG-T$_\mathrm{dwc}$. As in Tab.~\ref{tab: imagenet_con}, MSG-T$_\mathrm{dwc}$ shows promising performance with low FLOPs. We believe these newly proposed attention computing patterns will facilitate future vision Transformer research in various manners and scenarios.

\subsection{Analysis about Advantages of MSG Tokens}
We take Swin-~\cite{liu2021swin} and \name for comparison, and analyze their behaviors from two aspects as follows. 

\vspace{-5pt}
\paragraph{Receptive fields.} Let the window size be $W$. For Swin, the window is shifted by $\frac{W}{2}$ in every two Transformer blocks and the receptive field is $\left(\frac{3W}{2}\right)^2$ after two attention computations. For MSG, assuming the shuffle size is $S\ge2$, a larger receptive field of $\left(SW\right)^2$ is obtained with two attention computations.

\vspace{-5pt}
\paragraph{Information exchange.} In Swin, each patch token obtains information from other patch tokens in different windows, where valuable information is extracted by interacting with many other patch tokens. In MSG, information from one window is summarized by a MSG token and directly delivered to patch tokens in other windows. This manner eases the difficulty of patch tokens obtaining information from other locations and promotes the efficiency.

\begin{table}[t!]
\centering
\small
\caption{Ablation study about MSG token manipulation.}
\label{tab: abla_manip}
\begin{threeparttable}
\begin{tabular}{l|c|c|c}
\toprule
\textbf{Manip. Op.} & Shuffle & Average & Shift \\
\midrule
\textbf{ImageNet Top-1 (\%)} & \textbf{81.1} & 80.8 & 80.6 \\
\bottomrule
\end{tabular}
\end{threeparttable}
\end{table}

\subsection{Study about Manipulations on MSG Tokens}
As claimed in the main text, how to manipulate MSG tokens is not limited to the adopted shuffle operation. We study two additional manipulations, namely, the `average' (MSG tokens are averaged for the next-round attention) and `shift' (MSG tokens are spatially shifted). As in Tab.~\ref{tab: abla_manip}, `shuffle' works the best, and we conjecture that `average' lacks discrimination for different windows, and `shift' requires more stages to deliver information to all the other windows. We believe explorations on manipulation types carry great potential and will continue this as an important future work.

\end{document}